\documentclass{article}
\let\standardaddcontentsline\addcontentsline
\usepackage{iclr2027_conference,times}
\let\addcontentsline\standardaddcontentsline

\newif\ifanonymous
\anonymousfalse

\usepackage{amsmath,amsfonts,bm}

\def\eqref#1{equation~\ref{#1}}

\def\1{\bm{1}}

\DeclareMathAlphabet{\mathsfit}{\encodingdefault}{\sfdefault}{m}{sl}
\SetMathAlphabet{\mathsfit}{bold}{\encodingdefault}{\sfdefault}{bx}{n}

\usepackage{amsmath}
\usepackage{amssymb}
\usepackage{dsfont}
\usepackage{booktabs}
\usepackage{graphicx}
\usepackage{xspace}
\usepackage{xcolor}
\usepackage{colortbl}
\usepackage{wrapfig}
\usepackage{microtype}
\usepackage{hyperref}
\usepackage{url}
\usepackage{fontawesome5}
\usepackage{cleveref}

\DeclareFontFamily{U}{stix2bb}{}
\DeclareFontShape{U}{stix2bb}{m}{n} {<-> stix2-mathbb}{}

\usepackage{xcolor}
\usepackage{soul}
\usepackage{enumitem}
\sethlcolor{yellow}

\title{Rome Wasn’t Built in One Turn: Replay-Grounded Reward Signals for Long-Horizon Freeciv Agents}

\author{%
Yanchen Liu\textsuperscript{\dag,1} \quad
Xinyu Che\textsuperscript{\dag,2} \quad
Brihi Joshi\textsuperscript{\dag,1}\\
\textbf{Renzo Fukuda\textsuperscript{3} \quad
Junhua Liu\textsuperscript{1} \quad
Ruifeng Li\textsuperscript{1} \quad
Yunfei Ge\textsuperscript{4} \quad
Hang Yan\textsuperscript{2} \quad
Xiang Ren\textsuperscript{1}}\\
\textsuperscript{1}University of Southern California\\
\textsuperscript{2}Xi'an Jiaotong University\\
\textsuperscript{3}Stanford University\\
\textsuperscript{4}Northwestern University\\
\texttt{\{liuyanch,xiangren\}@usc.edu}
}

\ifanonymous
  \iclrfinalfalse
\else
  \iclrfinalcopy
\fi

\newcommand{\civharness}{\textsc{CivHarness}\xspace}
\newcommand{\scoreboard}{\textsc{scoreboard}\xspace}
\newcommand{\civtelescope}{\textsc{CivTelescope}\xspace}
\newcommand{\civmarshmallow}{\textsc{CivMarshmallow}\xspace}
\newcommand{\judge}{\textsc{judge}}
\newcommand{\sparse}{\textsc{sparse}\xspace}
\newcommand{\rem}[1]{\texttt{rem#1}}
\newcommand{\hto}{\raisebox{0.15ex}{$\scriptstyle\rightarrow$}\allowbreak}
\soulregister{\rem}{1}
\soulregister{\citep}{7}
\soulregister{\citet}{7}
\soulregister{\citealp}{7}
\soulregister{\civharness}{0}
\soulregister{\civtelescope}{0}
\soulregister{\civmarshmallow}{0}
\soulregister{\scoreboard}{0}
\soulregister{\sparse}{0}
\soulregister{\Cref}{7}
\soulregister{\ref}{7}
\soulregister{\hto}{0}
\newcommand{\phstart}[1][]{\textrm{phase}_{\textrm{start}}\if\relax\detokenize{#1}\relax\else^{\rem{#1}}\fi}
\newcommand{\phend}[1][]{\textrm{phase}_{\textrm{end}}\if\relax\detokenize{#1}\relax\else^{\rem{#1}}\fi}
\newcommand{\evalset}[1][]{\mathrm{E}\if\relax\detokenize{#1}\relax\else^{\rem{#1}}\fi}

\begin{document}

\maketitle
\ifanonymous\else
{\renewcommand{\thefootnote}{\fnsymbol{footnote}}\footnotetext[2]{Equal contribution.}}
\vspace{-18pt}
\begin{center}
\faGithub\ \textbf{Code}: \url{https://github.com/jamesnulliu/CivMarshmallow}\\
\raisebox{-0.15em}{\includegraphics[height=1em]{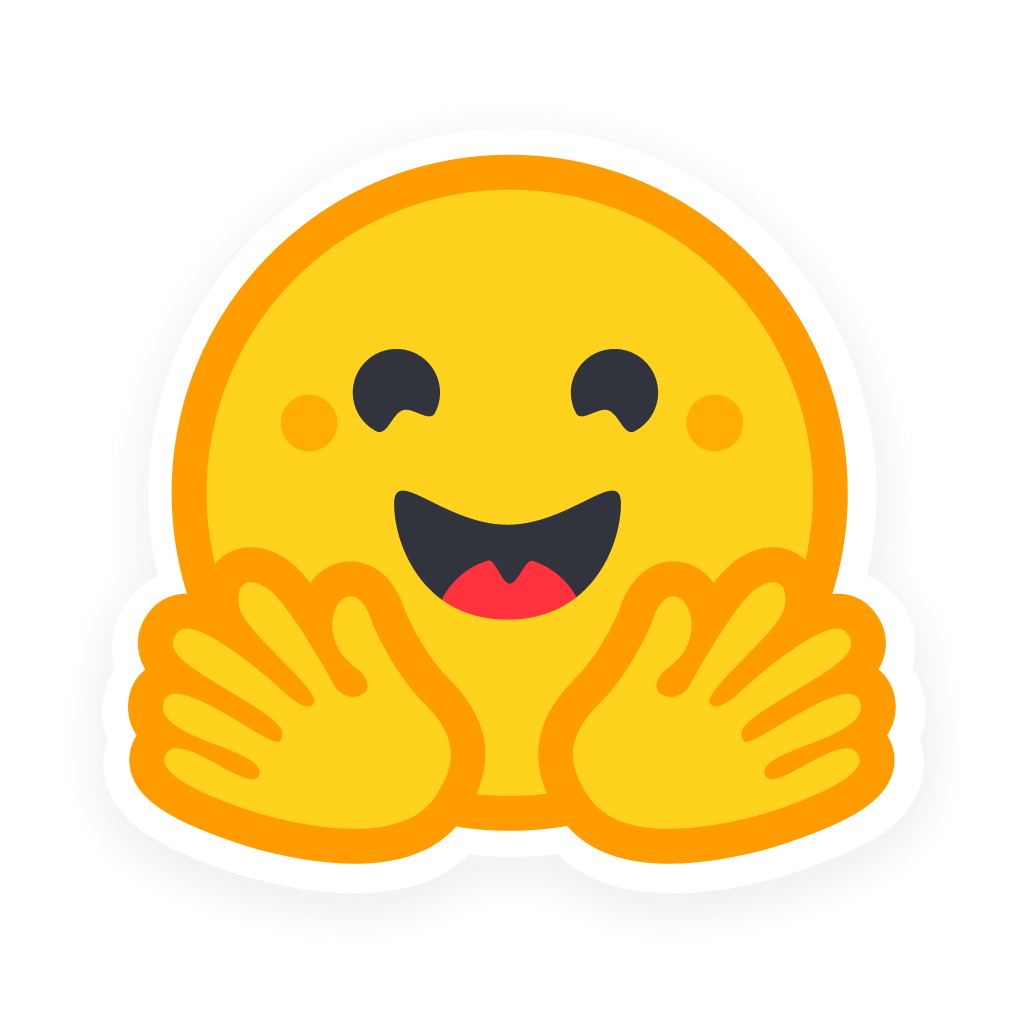}}\ \textbf{Checkpoints}: \href{https://huggingface.co/collections/CivMarshmallow/civmarshmallow}{\nolinkurl{CivMarshmallow/civmarshmallow}}
\end{center}
\vspace{2pt}
\lhead{}
\renewcommand{\headrulewidth}{0pt}
\fi

\addtocontents{toc}{\protect\setcounter{tocdepth}{-1}}

\begin{abstract}
Training language-model agents to play long-horizon strategy games is hard because deterministic feedback is sparse and often arrives only at the end of a game. 
In-game scores can offer dense signals, but they do not reliably predict future outcomes: in 8--68\% of position pairs across eight strategy games, the higher-scoring position leads to the worse outcome.
We term this phenomenon \textit{scoreboard traps}.
In Freeciv, an open-source turn-based empire-building game where several players manage units, cities, and diplomacy over hundreds of turns, we look into better reward signals for gameplay agents.
To facilitate this, we build \civharness{}, an open system that saves current game states, branches on different decisions, and replays the game to future turns to provide measurable long-horizon feedback from saved positions.
We use \civharness{} to train \civtelescope, an LLM Judge 
that predicts which one of two game positions will yield the better outcome.
On scoreboard-trap pairs, \civtelescope
is 1.3--2.0 times more accurate than off-the-shelf language models, and it generalizes to new game starts, later turns, and unseen rules.
We then train a Freeciv gameplay agent with \civtelescope as a dense reward, and call this recipe \civmarshmallow.
When used in the middle stage of the game where scoreboard traps are common, \civmarshmallow reaches an average training phase gain of 8.94 points, 77\% more than the best baseline, the end-of-game reward.
Over the full 120-turn game, \civmarshmallow uses a hybrid reward that starts from the scoreboard or the end-of-game reward and shifts to \civtelescope as the game advances; this hybrid reward scores highest among the rewards we compare.
\end{abstract}

\section{Introduction}

Large Language Model (LLM) agents have made substantial progress in decision-making in settings that require exploration and planning, like everyday web use \citep{zhou2024archer,feng2025gigpo}, mobile device control \citep{bai2025digiq}, and strategy game playing \citep{schultz2025mastering,belle2025agents,li2026saga}.
In strategy games, agents must learn to plan carefully, manage resources, and make tactical decisions.
Reinforcement learning (RL) with game-based feedback offers a way to train these LLM agents \citep{zhou2024archer,wang2025ragen,liu2025spiral}.
However, strategy games often involve \textit{long horizons}, with the consequences of decisions unfolding over many turns.
Agents must therefore learn from sparse, delayed feedback at the end of the game and connect these eventual outcomes to intermediate decisions.

\begin{figure}[t!]
\centering
\includegraphics[width=\linewidth]{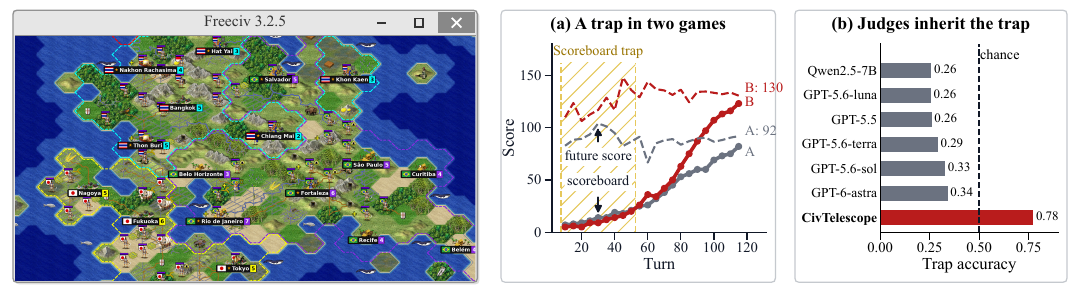}
\caption[Freeciv and the scoreboard trap.]{\textbf{Freeciv and the scoreboard trap.}
(a) Current scoreboard score (solid) and future score at turn 120 (dashed) in Freeciv games A (\textcolor[HTML]{6B7280}{gray}) and B (\textcolor[HTML]{B91C1C}{red}). A leads the scoreboard until turn 50, while B leads the future score during the trap (\textcolor[HTML]{997400}{yellow}).
(b) Frontier LLMs fail at identifying scoreboard traps given a pair of two positions.
}
\label{fig:headline}
\end{figure}

Outside games, process rewards score the intermediate steps of math solutions and agent trajectories~\citep{lightman2024verify,xia2025agentrm}; in games, one readily available source of dense feedback is the current in-game score, shown in the game's \textit{scoreboard}~\citep{mnih2015human,kuttler2020nethack,klissarov2024motif}.
However, this score may not reflect the long-term value of a current game position: the benefits of strategic investments often emerge later, while changing game conditions and other players’ actions can undermine an apparent lead. 
We call a pair of positions a \textit{scoreboard trap} when the position with the higher current score compared to the other leads to a worse future outcome (See \Cref{fig:headline}(a)).
Across eight strategy games with varying gameplay complexity, we demonstrate that such traps affect 8--68\% of position pairs (Figure~\ref{fig:xgame_traps}).

To study how to provide better reward signal for strategic gameplay agents, we focus our experiments on Freeciv, an open-source turn-based empire-building game~\citep{freeciv2026}.
We introduce \civharness{} (\Cref{sec:harness}), a reusable harness for agentic reinforcement learning that saves game states, branches on alternative decisions, and replays their continuations to provide training feedback.
With \civharness{}, we find that frontier LLMs rarely detect scoreboard traps: when asked which of two positions leads to the better future outcome, five frontier models reach only 26\%--34\% accuracy on held-out trap pairs, below chance (\Cref{fig:headline}(b)).
We instead train \civtelescope (\Cref{sec:judge}), a pairwise language-model judge, on future outcomes averaged over replays from each position, which reaches 77.5\% accuracy on the same pairs.

We use \civtelescope to provide dense rewards for training a Freeciv gameplay agent (\Cref{sec:signals}), with the final game score as the training objective, and call this recipe \civmarshmallow\footnote{As in the marshmallow test of delayed gratification \citep{mischel1972cognitive}, \civtelescope rewards a position for an outcome that arrives many turns later, even when the position gives up score now.}.
Training proceeds in four phases, with 20, 40, 80, and 120 turns remaining, allowing us to compare \civmarshmallow with sparse end-of-game and dense scoreboard rewards at each phase.
With 80 turns remaining, where scoreboard traps are common and \civtelescope can already rank positions, \civmarshmallow improves the phase gain by 77\% over the best baseline (the end-of-game reward).
For the full 120-turn game, \civmarshmallow shifts from the scoreboard or the end-of-game outcome to \civtelescope as the game advances, which keeps the early-game signal stable; this hybrid reward scores highest among the rewards we compare.

Our contributions are:
\begin{enumerate}[nosep]
    \item \textbf{Scoreboard traps.} We show that scoreboard traps are common across eight strategy games and that frontier LLMs fall below chance on them in Freeciv (\Cref{sec:exp_trap}).
    \item \textbf{\civharness{}.} We release an open-source harness for agentic RL in Freeciv that saves, branches, and replays game states, so any feedback signal can be checked against actual future outcomes (\Cref{sec:harness}).
    \item \textbf{\civtelescope.} We train a pairwise judge on replay outcomes that generalizes to new starts, later turns, and an unseen ruleset; trained on a frontier model's preferences instead, the same judge stays below chance (\Cref{sec:judge,sec:exp_repair}).
    \item \textbf{\civmarshmallow.} We train Freeciv agents in phases with \civtelescope as a dense reward, alone and combined with scoreboard or end-of-game rewards over the full game (\Cref{sec:signals,sec:exp_phase,sec:exp_hybrid}).
\end{enumerate}

\section{Related Work}
\label{sec:related}

\paragraph{Turn-based strategy-game agents.}
CivRealm provides a Freeciv environment for learning and language-based agents \citep{qi2024civrealm}; SAGA and Vox Deorum add planning structure \citep{li2026saga,chen2025voxdeorum}, and Civilization benchmarks report sparse terminal feedback over hundreds of turns \citep{chen2026civbench,andrews2026civbench}.
These environments start each run from a new game or a designed scenario, whereas \civharness{} branches and replays any saved position, so policies and rewards can be compared against future outcomes from shared positions.
Unlike the value estimator of \citet{chen2026civbench} which learned from played games to evaluate agents, \civtelescope learns from replayed branches of the same position and serves as a training reward.

\paragraph{Model feedback and proxy rewards.}
Language- and vision-model feedback can be distilled into rewards \citep{klissarov2024motif,wang2024rlvlmf,cao2024dense,xie2024text2reward}, but it depends on the model recognizing progress; optimizing a proxy can diverge from the intended objective \citep{gao2023scaling,razin2025teacher}, and game benchmarks document gaps between visible scores and progress \citep{kuttler2020nethack,paglieri2025balrog}.
We measure these gaps against replayed outcomes and show that replay labels, unlike preference labels, do not inherit the labeler's trap errors.

\paragraph{Rollout-grounded feedback and curricula.}
Outside games, process supervision and value estimation use step labels, Monte-Carlo rollouts, or progress measures \citep{lightman2024verify,wang2024mathshepherd,yuan2025free,setlur2025rewarding,kazemnejad2025vineppo}, and agent methods learn rewards or values from interaction \citep{xia2025agentrm,bai2025digiq,zhou2024archer,feng2025gigpo}.
We instead train a pairwise judge from rollout outcomes in a multiplayer strategy game and calibrate it with reference positions of known value.
Following reverse curricula \citep{florensa2017reverse,salimans2018improving}, we increase the remaining turns from phase to phase and compare reward signals within each phase.

\section{Background}
\begin{figure}[t]
\centering
\includegraphics[width=\linewidth]{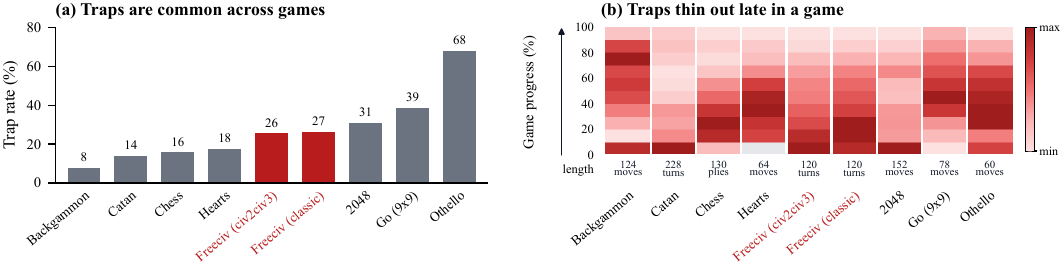}
\caption[Scoreboard traps.]{\textbf{Scoreboard traps.}
(a) Trap rate across eight strategy games.
(b) Trap rate by game progress; darker red is a higher rate, scaled within each game.
}
\label{fig:xgame_traps}
\end{figure}

\paragraph{Freeciv.}

Freeciv~\citep{freeciv2026} is an open-source turn-based strategy game modeled on Sid Meier's Civilization\footnote{\url{https://civilization.2k.com/}}.
Each player leads a civilization from its first settlers: it founds and grows cities, researches technologies, builds units and wonders, and negotiates or fights with the other players.
In our games, the agent controls one of four players on a 1000-tile map; the other three are Freeciv's built-in AI players at a fixed \textit{bot difficulty} (\Cref{sec:harness}).
The agent sees only what its units and cities have explored.
Freeciv keeps a running score for every player, computed from quantities such as citizens, technologies, wonders, and culture, and shows it on the in-game scoreboard; we call the agent's score at the end of the game its \textit{terminal score}.
The games follow one of two rulesets, \textsc{civ2civ3} (the Freeciv 3 default, used in our main experiments) or \textsc{classic} (Appendix~\ref{app:rulesets}).

\paragraph{Scoreboard traps across strategy games.}
\label{sec:exp_trap}

Strategy games often suffer from \textit{scoreboard traps}: when the higher current position can lead to a worse future outcome.
To measure the traps, we build pairs of positions from recorded games and count the fraction of pairs in which the scoreboard order opposes the order of their future outcomes.
Scoreboard traps occur in 8--68\% of decidable position pairs across eight games (\Cref{fig:xgame_traps}(a)), including 26\% in Freeciv under \textsc{civ2civ3} and 27\% under \textsc{classic} (Appendix Table~\ref{tab:oracle_specs}).
Trap rates also vary with game progress (\Cref{fig:xgame_traps}(b)): in \textsc{civ2civ3}, the rate falls from 55\% in the first tenth of the game to 8\% in the last (Appendix~\ref{app:xgame}).
In Freeciv, the scoreboard also becomes a worse predictor as games get longer: from 60-turn to 80-turn games, its correlation with the terminal score falls from 0.81 to 0.12 and the trap rate rises from 30\% to 54\% (Appendix Table~\ref{tab:trap_horizon}).

\paragraph{Off-the-shelf LLM judges.}
On held-out Freeciv trap pairs, all six LLMs, from frontier closed models to a small open model, score below chance (0.26--0.34), even though they reach 0.77--0.82 on non-trap pairs (\Cref{fig:headline}(b); Appendix Table~\ref{tab:trap_accuracy}).
Warning them in the prompt that visible scores can mislead does not help: the frontier models still rank only 0.32--0.38 of trap pairs correctly.
The failure also appears in simpler games: four of five frontier models rank below chance in Chess, and two of five in Othello (Appendix Table~\ref{tab:xgame_judges}).

\section{Designing and Building \civharness{}}
\label{sec:harness}

To facilitate our training and evaluation, we introduce \civharness, a reusable harness for agentic reinforcement learning in Freeciv.
From a saved state, \civharness runs $K$ deterministically seeded branches on the real Freeciv game engine, either to the end of the game or for a chosen number of turns, and records their observations, actions, and outcomes; replay checks verify that the same save, seed, and action trace reproduce the same state.

A gameplaying agent 
interacts through configurable observations and actions, including city production, research, and unit orders executed by the game engine.
The agent sees only its own player's view of the game, not the full game state.
We also speed gameplay rollouts by implementing batch execution, which manages independent games, retries on failed gameplay runs, and records completed work so collection can resume.

\civharness records the agent's ordinary rollouts, and restoring a saved position lets it collect sibling rollouts that take different decisions from the same state, for group-based training and for comparing reward signals from shared starts.
Game logs report observations, issued actions, branch seeds, and bot difficulty.
Researchers using \civharness can therefore reuse position banks, replace a policy or reward, and reproduce the resulting comparisons with the same replay checks.

\begin{figure}[t!]
\centering
\includegraphics[width=1.0\linewidth]{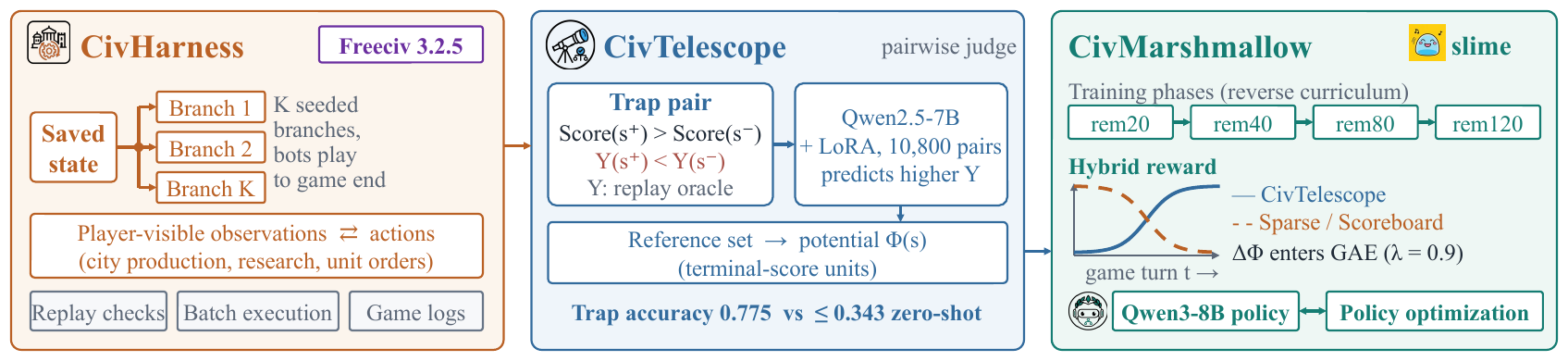}
\caption[Overview of the harness, judge, and training recipe.]{\textbf{Overview of \civharness, \civtelescope, and \civmarshmallow.}
\civharness branches saved Freeciv states and exchanges player-visible observations and actions with the agent.
\civtelescope, a pairwise judge trained on replay-labeled trap pairs, gives a potential $\Phi(s)$ through a reference set.
\civmarshmallow trains the policy in four phases with \civtelescope reward, and with a hybrid reward for full games.}

\label{fig:civharness_workflow}
\end{figure}

\Cref{fig:civharness_workflow} shows how \civharness connects to slime, the reinforcement-learning training engine used in our experiments.
The rollout interface exchanges player-visible observations and actions between the policy and the game engine, converts \civtelescope{}'s pairwise preferences into state values with a reference set for judge-based rewards (\Cref{eq:potential}), and passes the resulting samples to slime, which updates the policy.
Appendix~\ref{app:harness} provides further details on the harness design.

\section{Training a Freeciv gameplaying agent}
\label{sec:method}

In this section, we describe using \civharness to first create \textit{trap pairs} to train an LLM judge, \civtelescope, that can escape the scoreboard trap.
We then detail \civmarshmallow, our recipe for training a Freeciv gameplay agent with \civtelescope rewards, used alone or combined with other rewards over the course of a game.

\subsection{Replay oracle and trap pairs}
\label{sec:oracle}

We define a \textit{replay oracle} as a mechanism that takes the current game state and uses \civharness to observe future game states, possible till the end of the game.
For a game state $s$, the replay oracle creates $K$ branches forward from $s$, all the way to the end of the game with Freeciv's built-in bots at a fixed bot difficulty.
We define $Y(s)$ as the player's mean terminal score over these branches from $s$.
For positions taken from policy rollouts, $Y(s)$ is the realized terminal score of that rollout.
When the agent's policy $\pi$ plays the player instead of the bots, we write $Y_\pi(s)$ for its mean terminal score from $s$.

We use the replay oracle to create \textit{trap pairs} that explicitly capture the scoreboard trap, as demonstrated in~\Cref{sec:exp_trap}.
A pair $(s^+,s^-)$ is a trap pair when
\begin{equation}
    \scoreboard(s^+) > \scoreboard(s^-), \qquad Y(s^+) < Y(s^-).
\end{equation}
We control the starting position, replay length, branch count, and bot difficulty when generating branches.
Since the terminal score is not the \textit{true} terminal score but rather future replays, we only pick pairs where the replay oracle's terminal scores differ more than twice their pooled error. 
This is done so that only significantly different states are considered.
Pairs of rollouts that share the same start and are compared at the same turn show the trap at equal or higher rates (Appendix~\ref{app:sibling}).

\subsection{Using \civharness to build \civtelescope}
\label{sec:judge}

\civtelescope is a pairwise language-model-based judge, fine-tuned from Qwen2.5-7B. It predicts which position has the higher terminal score under the replay oracle.
The input of \civtelescope is a pair of game positions, each rendered as text from the evaluated player's point of view: every player's score together with that player's gold, technologies, cities, and units.
The output is the position expected to reach the higher terminal score; because the answer can depend on which position is listed first, we query both orders and average the two answers.
The training labels are the terminal scores of two kinds of games: branches played to the end by Freeciv's bots from saved positions, which give $Y(s)$, and the agent's own rollouts.
A training pair holds two positions from different games at most ten turns apart whose labels differ by more than twice their pooled standard error, and we fine-tune the model with LoRA (rank 16) on 10,800 such pairs.
Section~\ref{sec:signals} turns its pairwise preferences into a reward.

\subsection{\civmarshmallow: Training recipe for Freeciv gameplay agents}
\label{sec:signals}

We now provide the full training recipe, \civmarshmallow, for training a Freeciv gameplaying agent. 
The recipe has three parts: a phased training schedule that moves from late-game to full-game starts, the \civtelescope reward, which turns the judge's preferences into a dense reward, and, for full games, a hybrid reward that combines \civtelescope with a sparse or scoreboard reward.
We train the baseline rewards, \sparse and \scoreboard, with the same schedule, so that every reward is compared under the same conditions.

\paragraph{Phased training approach.}
Every gameplay agent, whether trained with \civmarshmallow or with a baseline reward, is trained on late-game positions first and on earlier positions later.
We denote the number of game turns remaining to play by $\rem{}$, and the maximum number of turns in our gameplay is 120.
The gameplay agent is trained sequentially in four \emph{training phases}, \rem{20}, \rem{40}, \rem{80}, and \rem{120}, each with more turns remaining than the one before.
Each training phase starts from the policy checkpoint of the preceding phase and a fixed position bank 
, a set of saved game states from which training games start (turn 100, 80, 40, and 1).
Training thus follows a reverse curriculum, which starts near the end of a task and moves the start earlier as training proceeds \citep{florensa2017reverse, salimans2018improving}.

\paragraph{Rewards.}
We compare two baseline rewards, \sparse and \scoreboard, with the \civtelescope reward of \civmarshmallow; all three use the same rollout budget.
\sparse rewards use only the terminal score, which is only available at the end of gameplay.
Its advantage normalizes the terminal score across the rollouts from the same starting position, as in GRPO \citep{shao2024deepseekmath}.
\scoreboard rewards use the current scoreboard value from Freeciv as a dense reward signal.
In order to use \civtelescope as a reward, we convert
its pairwise preferences into a scalar potential with a \emph{reference set} $\mathcal{B}=\{r_j\}_{j=1}^{M}$: positions $r_j$ from nearby game turns whose terminal scores $Y(r_j)$ are known.
Let $p_\theta(s\succ r)$ be the judge's probability that $s$ leads to the better outcome.
The potential is the reference score at the quantile given by the mean win probability,
\begin{equation}
    \Phi_{\mathcal{B}}(s)=Q_{\mathcal{B}}\Bigl(\tfrac{1}{M}\textstyle\sum_{j=1}^{M}p_\theta(s\succ r_j)\Bigr),
\label{eq:potential}
\end{equation}
where $Q_{\mathcal{B}}(w)$ linearly interpolates the sorted $Y(r_j)$ at quantile $w$.
The judge supplies the ranking and the reference set supplies the scale, so $\Phi$ has the units of $R$, the realized terminal score of a rollout.
A fixed reference set saturates as the policy improves, so the set evolves with the policy.
Every $N$ policy updates we form a candidate $\tilde{\mathcal{B}}$ from the most recent training rollouts, taking positions whose terminal scores are evenly spaced in rank and keeping the strongest bot-played positions, and accept it only if its potential still ranks held-out recent rollouts:
\begin{equation}
    \mathcal{B}^{(n)}=
    \begin{cases}
        \tilde{\mathcal{B}}^{(n)} & \text{if } N\mid n \text{ and } \rho\bigl(\Phi_{\tilde{\mathcal{B}}^{(n)}},R\bigr)\geq\rho_{\min},\\
        \mathcal{B}^{(n-1)} & \text{otherwise,}
    \end{cases}
\label{eq:reference_update}
\end{equation}
where $\rho$ is the Spearman correlation between the potential and the terminal score on those rollouts.
Appendix~\ref{app:judge} gives $M$, $N$, $\rho_{\min}$, and the construction for each game turn.

For a state sequence $s_0,\ldots,s_T$, potential differences enter generalized advantage estimation \citep{schulman2016gae, ng1999policy}.
With $\gamma=1$ and $\lambda<1$,
\begin{equation}
 A_t = \sum_{k=t}^{T-1}\lambda^{k-t}\bigl(\Phi(s_{k+1})-\Phi(s_k)\bigr)
       + \lambda^{T-t}\bigl(R-\Phi(s_T)\bigr).
\label{eq:gae}
\end{equation}
Using $\lambda<1$ retains information from intermediate potential differences, while the observed terminal score remains the final target.
With $\lambda=1$, the differences cancel and the estimator reduces to $A_t=R-\Phi(s_t)$, in which the potential serves only as a baseline \citep{schulman2016gae}.
Section~\ref{sec:exp_ablation} ablates this choice with a variant that uses \civtelescope only on the final position of each game.

\paragraph{Combining rewards to produce a hybrid reward for training.}
\label{sec:hybrid}
For full games, to keep the early-game signal stable, \civmarshmallow combines the \civtelescope reward with a baseline reward, using the baseline early in the game and \civtelescope later (\Cref{fig:civharness_workflow} right, Section~\ref{sec:exp_hybrid}); we call these the hybrid rewards, \scoreboard{}\hto\civtelescope and \sparse{}\hto\civtelescope.
At \rem{120}, a hybrid reward increases the weight of \civtelescope as each game advances.
The weight at game turn $u$ is $w(u)=(1+\exp(-(u-\tau)/\beta))^{-1}$, with transition turn $\tau$ and width $\beta>0$; $u_t$ denotes the game turn of $s_t$.
Each rollout uses this schedule independently of the training phases.

\scoreboard{}\hto\civtelescope combines potentials on a common scale.
We map the current score's rank among the reference positions to their terminal-reward scale, giving a calibrated scoreboard potential $\Phi_{\mathrm{sb,cal}}$.
We then use
\begin{equation}
    \Phi_{\mathrm{hybrid}}(s_t)=(1-w(u_t))\Phi_{\mathrm{sb,cal}}(s_t)+w(u_t)\Phi_{\mathrm{judge}}(s_t)
\label{eq:hybrid_potential}
\end{equation}
in Equation~\ref{eq:gae}.
\sparse{}\hto\civtelescope combines the group-normalized terminal advantage $A^{\mathrm{sparse}}$ with the judge advantage $A_t^{\mathrm{judge}}$ from Equation~\ref{eq:gae} as
\begin{equation}
    A_t^{\mathrm{hybrid}}=(1-w(u_t))A^{\mathrm{sparse}}+w(u_t)A_t^{\mathrm{judge}}.
\label{eq:hybrid_advantage}
\end{equation}
The reverse order, \civtelescope{}\hto\sparse, replaces $w(u_t)$ with $1-w(u_t)$.
\scoreboard{}\hto\civtelescope{}\hto\sparse uses \scoreboard early in the game, \civtelescope in the middle, and \sparse from turn 80 (parameters in Appendix~\ref{app:hybrid}).

\section{Experiments}
\label{sec:experiments}

\subsection{Setup}

\paragraph{Model and Training Parameters.}

The gameplay agent is Qwen3-8B \citep{yang2025qwen3} with thinking disabled, and \civtelescope is fine-tuned from Qwen2.5-7B-Instruct \citep{qwen2025qwen25}.
All reward families update every parameter of the agent with the same clipped policy objective, without KL or entropy terms, and differ only in the advantage (Section~\ref{sec:signals}).
Each training phase runs 12 iterations, and each iteration plays eight rollouts from each of eight starting positions drawn from the phase's position bank.
We use Adam with a constant learning rate of $10^{-6}$, clipping ratios of 0.2 below and 0.28 above, sampling temperature 1.0, and $\lambda=0.9$ in Equation~\ref{eq:gae}.

\paragraph{Evaluation.}
We construct the evaluation sets from 45 held-out games that are unseen during training: each training phase \rem{X} has its own evaluation set $\evalset[X]$, the 45 positions at the turn at which that phase begins, each played to the end of the game.
A gameplay agent checkpoint plays on $\evalset$, the evaluation set of the current phase, before and after each training phase (from the $\phstart$ to the $\phend$ checkpoint), eight games per position.
Its \emph{terminal score} $S=\tfrac{1}{|\evalset|}\sum_{p\in\evalset}Y_\pi(p)$ averages $Y_\pi(p)$, the agent's mean terminal score over the eight games from position $p$ (Section~\ref{sec:oracle}).
The \emph{phase gain} $G=S_{\phend}-S_{\phstart}$ is the change in $S$ over a training phase, and $g_p$ is that change at position $p$ alone.
The \emph{paired win rate} $W_{a,b}=\tfrac{1}{|\evalset|}\sum_{p\in\evalset}\mathds{1}[g_p^{a}>g_p^{b}]$ is the fraction of positions on which method $a$ gains more than method $b$.
In this work, we primarily care about $G$.
\civtelescope quality is measured separately by accuracy on held-out trap pairs.
In order to mitigate position bias \citep{zheng2023judging,wang2024fair},
we query the judge twice on every pair, once with each position listed first, and average the two results.

\subsection{\civtelescope Results}
\label{sec:exp_repair}

\begin{wraptable}[20]{r}{0.5\textwidth}
\setlength{\abovecaptionskip}{0pt}
\vspace{-10pt}
\caption[Trap accuracy of judges.]{Trap accuracy on held-out games (in-domain) and an unseen ruleset; chance is 0.5.
$\uparrow$: higher is better. $^\dag$No significant difference between the two marked values.}
\label{tab:judge_comparison}
\begin{center}
\footnotesize
\setlength{\tabcolsep}{3pt}
\renewcommand{\arraystretch}{1.0}
\begin{tabular*}{\linewidth}{@{\extracolsep{\fill}}lrr@{}}
\toprule
Method & \begin{tabular}[b]{@{}r@{}}Held-out\\games $\uparrow$\end{tabular} & \begin{tabular}[b]{@{}r@{}}Unseen\\ruleset $\uparrow$\end{tabular} \\
\midrule
\rowcolor{black!8}
\multicolumn{3}{@{}l}{\textcolor{black!65}{\textbf{Baselines}}} \\
\addlinespace[1pt]
\textbf{\scoreboard} & 0 & 0 \\
\textbf{7B backbone}, zero-shot & 0.260 & 0.241 \\
\textbf{GPT-5.6-luna}, zero-shot & 0.258 & 0.203 \\
\textbf{GPT-5.5}, zero-shot & 0.262 & 0.285 \\
\textbf{GPT-5.6-terra}, zero-shot & 0.293 & 0.268 \\
\textbf{GPT-5.6-sol}, zero-shot & 0.330 & 0.250 \\
\textbf{GPT-6-astra}, zero-shot & 0.343 & 0.488 \\
\textbf{Preference-labeled judge} (7B) & 0.391 & 0.415 \\
\textbf{Regression value head} (7B) & 0.520 & \textbf{0.622}$^\dag$ \\
\midrule
\rowcolor{black!8}
\multicolumn{3}{@{}l}{\textcolor{black!65}{\textbf{Replay-grounded judge}}} \\
\addlinespace[1pt]
\textbf{\civtelescope} & \textbf{0.775} & \textbf{0.592}$^\dag$ \\
\bottomrule
\end{tabular*}
\end{center}
\end{wraptable}

\paragraph{\civtelescope vs other judges.}\leavevmode %
\Cref{tab:judge_comparison} compares \civtelescope with the scoreboard, zero-shot open and frontier models, a Qwen2.5-7B-Instruct judge distilled from the preferences of GPT-5.6-sol \citep{openai2026gpt56}, and a regression value head on the same backbone, as in gameplay RL \citep{silver2016mastering,silver2018general}, on held-out games and on an unseen ruleset (trained on civ2civ3, tested on classic).
On held-out games, \civtelescope ranks 77.5\% of trap pairs correctly, where its own zero-shot backbone and five frontier models rank 26--34\% (Figure~\ref{fig:headline}b).
It is also the most accurate on non-trap pairs, 0.878 against 0.77--0.82, so it does not simply reverse the score order (Appendix Table~\ref{tab:trap_accuracy}).
A regression value head trained on the same positions with replay outcomes as targets is at chance on held-out trap pairs (0.520, against 0.688 for \civtelescope on the same pairs; Appendix~\ref{app:value_head}); on the unseen ruleset the two are not distinguishable on trap pairs (0.622 and 0.592, overlapping intervals), and \civtelescope stays ahead on non-trap pairs (0.741 against 0.703).
Out of domain with unseen rules, \civtelescope reaches 0.592 on classic trap pairs, against 0.241 for its backbone and 0.20--0.49 for five frontier models.

\subsection{\civmarshmallow vs. baseline rewards across training phases}
\label{sec:exp_phase}
Table~\ref{tab:phase_diagram} reports the held-out phase gain $G$ of the two baseline rewards and of \civmarshmallow with the \civtelescope reward in each training phase.
\civtelescope alone helps most in the middle game: at \rem{80} it gains 8.94 points, against
5.04 for \sparse and 0.83 for \scoreboard,
and its paired win rate $W$ is 81\% against sparse and 96\% against scoreboard, with both differences positive in every seed (Table~\ref{tab:phase_diagram}).
\scoreboard is a good proxy for the terminal score towards the end of the game (\rem{20}) or right at the beginning (\rem{120}), whereas \sparse rewards are good towards the end of the game as well, at \rem{40}.

\begin{table}[t]
\caption[Held-out phase gain in each training phase.]{%
Held-out phase gain $G$ ($\uparrow$: higher is better) in each training phase: mean over three training seeds, with per-seed values in parentheses; each value averages the 45 positions of the evaluation set.
Bold marks the best mean in each phase.}
\label{tab:phase_diagram}
\begin{center}
\footnotesize
\setlength{\tabcolsep}{3pt}
\renewcommand{\arraystretch}{1.08}
\begin{tabular*}{\linewidth}{@{\extracolsep{\fill}}lrrrr@{}}
\toprule
Reward & \rem{20} & \rem{40} & \rem{80} & \rem{120} \\
\midrule
\rowcolor{black!8}\multicolumn{5}{@{}l}{\textcolor{black!65}{\textbf{Baselines}}} \\
\sparse & 0.53\,{\scriptsize(1.1/0.9/$-$0.4)} & \textbf{7.11}\,{\scriptsize(7.2/7.0/7.1)} & 5.04\,{\scriptsize(4.7/3.0/7.4)} & 2.87\,{\scriptsize(4.3/4.6/$-$0.3)} \\
\scoreboard & \textbf{2.72}\,{\scriptsize(1.4/3.3/3.5)} & 3.68\,{\scriptsize(7.0/1.9/2.2)} & 0.83\,{\scriptsize($-$0.1/1.4/1.1)} & 4.87\,{\scriptsize(5.8/6.3/2.5)} \\
\rowcolor{black!8}\multicolumn{5}{@{}l}{\textcolor{black!65}{\textbf{\civmarshmallow (Ours)}}} \\
\civtelescope & 0.62\,{\scriptsize(1.9/$-$0.3/0.3)} & 3.61\,{\scriptsize(7.4/3.9/$-$0.4)} & \textbf{8.94}\,{\scriptsize(5.4/9.9/11.6)} & 2.77\,{\scriptsize(5.1/2.6/0.6)} \\
\scoreboard{}\hto\civtelescope & -- & -- & -- & \textbf{5.21}\,{\scriptsize(6.2/5.5/3.9)} \\
\sparse{}\hto\civtelescope & -- & -- & -- & 5.15\,{\scriptsize(7.2/5.2/3.1)} \\
\bottomrule
\end{tabular*}
\end{center}
\end{table}

We note that improvements from \civtelescope are most prominent where scoreboard traps are common (Appendix Table~\ref{tab:knowability_spectrum}).
In the policy's own 120-turn training games, rollouts from the same start are scoreboard traps 18--22\% of the time over turns 30--69, where \rem{80} begins, whereas \rem{40} and \rem{20} begin after the rate has fallen to 10\% and 7\% (Appendix Table~\ref{tab:sibling_trap}).
\civtelescope thus gives the best reward where traps are common, which motivates the hybrid rewards of \civmarshmallow for full games: a baseline reward early in the game and \civtelescope later.

\subsection[\civmarshmallow with hybrid rewards for full games]{\civmarshmallow with hybrid rewards for full games}
\label{sec:exp_hybrid}

Hybrid rewards gain the most over the full game (\rem{120}): \scoreboard{}\hto\civtelescope gains 5.21 points and \sparse{}\hto\civtelescope 5.15, against 4.87 for \scoreboard, 2.87 for \sparse, and 2.77 for \civtelescope alone (Table~\ref{tab:phase_diagram}).
Only pure \civtelescope starts from the same checkpoint as the hybrids, and both hybrids beat it in all three seeds, by 2.4 points on average.

Early in the game the score tracks growth, while \civtelescope cannot yet tell positions apart: in turns 10--29 its potential has a rank correlation of 0.09--0.10 with the terminal score, against 0.34--0.36 for the scoreboard (Appendix Table~\ref{tab:knowability_spectrum}).
From turn 30 on its correlation rises to 0.41--0.60 while traps remain common (Appendix Table~\ref{tab:sibling_trap}), so the hybrid rewards hand over to \civtelescope at turn 35.

\paragraph[\civmarshmallow vs. prompting agents.]{\civmarshmallow vs. prompting agents.}
Table~\ref{tab:scaffold_panel} compares the terminal score $S$ of the trained policies with five prompting-based agents re-implemented for our interface (Appendix~\ref{app:scaffolds}); all agents use Qwen3-8B \citep{yang2025qwen3} and the same action interface.
Every trained policy outscores every prompting agent: the strongest \civmarshmallow policy, trained with the hybrid reward \sparse{}\hto\civtelescope, reaches 23.4 against 13.2 for the best scaffold, and it is eliminated in 4\% of games against 9--26\%.
The trained policies also almost never propose an illegal action that is not allowed in the gameplay (0.1--0.5\% of decisions), where the scaffolds reach 41.6\%.

\begin{table}[t]
\setlength{\abovecaptionskip}{0pt}
\caption[Comparison with prompting agents.]{%
Terminal score $S$ and gameplay statistics on the 45 held-out full-game positions (120 turns, eight games per position); trained policies are evaluated after the final phase (\rem{120}).
\emph{Eliminated}: \% of games in which the agent's player is destroyed; \emph{invalid actions}: \% of decisions rejected by the interface.
$\uparrow$/$\downarrow$: higher/lower is better.}
\label{tab:scaffold_panel}
\begin{center}
\small
\setlength{\tabcolsep}{5pt}
\renewcommand{\arraystretch}{1.08}
\begin{tabular*}{\linewidth}{@{\extracolsep{\fill}}lrrrr@{}}
\toprule
Method & \shortstack[r]{Terminal\\score $S$ $\uparrow$} & \shortstack[r]{Eliminated\\(\%) $\downarrow$} & \shortstack[r]{Invalid actions\\(\%) $\downarrow$} & \shortstack[r]{Calls per\\decision} \\
\midrule
\rowcolor{black!8}
\multicolumn{5}{@{}l}{\textcolor{black!65}{\textbf{Prompting agents}}} \\
\addlinespace[2pt]
\textbf{Base model (Qwen3-8B)} & 7.38 & 30.0 & 6.0 & 1.00 \\
\textbf{Direct prompting} & 7.62 & 26.2 & 5.7 & 1.00 \\
\textbf{BaseLang} \citep{qi2024civrealm} & 11.65 & 12.4 & 16.2 & 1.00 \\
\textbf{Mastaba} \citep{qi2024civrealm} & 11.30 & 16.2 & 24.7 & 4.00 \\
\textbf{SAGA} \citep{li2026saga} & 13.23 & 8.8 & 41.6 & 1.10 \\
\textbf{Reflexion} \citep{shinn2023reflexion} & 10.22 & 15.1 & 36.7 & 1.01 \\
\midrule
\rowcolor{black!8}
\multicolumn{5}{@{}l}{\textcolor{black!65}{\textbf{RL baselines}}} \\
\addlinespace[2pt]
\textbf{\sparse} & 23.21 & 4.2 & 0.3 & 1.00 \\
\textbf{\scoreboard} & 23.29 & \textbf{3.5} & 0.3 & 1.00 \\
\midrule
\rowcolor{black!8}
\multicolumn{5}{@{}l}{\textcolor{black!65}{\textbf{\civmarshmallow (Ours)}}} \\
\addlinespace[2pt]
\textbf{\civtelescope} & 21.30 & 9.7 & \textbf{0.1} & 1.00 \\
\textbf{\scoreboard{}\hto\civtelescope} & 22.37 & 6.0 & 0.5 & 1.00 \\
\textbf{\sparse{}\hto\civtelescope} & \textbf{23.35} & 4.0 & 0.4 & 1.00 \\
\bottomrule
\end{tabular*}
\end{center}
\end{table}

\subsection{Ablations}
\label{sec:exp_ablation}

\begin{table}[t]
\setlength{\abovecaptionskip}{0pt}
\caption[Ablations.]{Ablations.
Each block changes one component and keeps the rest fixed.
Trap accuracy is evaluated on the 600 held-out trap pairs. 
$\uparrow$: higher is better.}
\label{tab:ablation}
\begin{center}
\small
\begin{tabular*}{\linewidth}{@{\extracolsep{\fill}}lllr@{}}
\toprule
Component & Variant & Metric & Result \\
\midrule
Label source & Replay outcomes (ours) & Trap accuracy $\uparrow$ & 0.610 \\
 & GPT-5.6-sol preferences & & 0.391 \\
\midrule
Judge input & Full state (ours) & Trap accuracy $\uparrow$ & 0.778 \\
 & Score numbers masked & & 0.687 \\
 & Score numbers only & & 0.113 \\
\midrule
Hybrid order & \scoreboard{}\hto\civtelescope (ours) & Phase gain $G$ at \rem{120} $\uparrow$ & 5.21 \\
 & \sparse{}\hto\civtelescope (ours) & & 5.15 \\
 & \scoreboard{}\hto\civtelescope{}\hto\sparse & & 2.84 \\
 & \civtelescope{}\hto\sparse & & 1.76 \\
\bottomrule
\end{tabular*}
\end{center}
\end{table}

\paragraph{Label source.}
Replacing the replay oracle's rewards with a frontier model's preferences brings the trap back.
We train two judges that are identical except for their labels: one learns from replay outcomes, the other from GPT-5.6-sol's preferences on the same pairs, as in Motif \citep{klissarov2024motif}.
The preference-distilled judge reaches 0.391 on held-out trap pairs, near its labeler's own 0.330, whereas the replay-labeled judge reaches 0.610 (Table~\ref{tab:ablation}).

\paragraph{Inputs provided to \civtelescope.}
As mentioned in Section~\ref{sec:judge}, \civtelescope's input is a text description of a position: every player's score together with the evaluated player's gold, technologies, cities, and units.
With every score number masked, trap accuracy is 0.687, where the zero-shot backbone reaches 0.275 on the full input; with only the score numbers left and without textual description, it falls to 0.113 while non-trap accuracy stays at 0.842 (Table~\ref{tab:ablation}).
A score-only judge is right on non-trap pairs and wrong on trap pairs, which is the scoreboard heuristic itself; the rest of the position is what lets \civtelescope escape it.
Therefore, \civtelescope does use the visible scores but does not reduce to them.

\paragraph{Using \civtelescope as a reward.}
Applying \civtelescope only to the final position of a game, instead of at every stage of the game, removes its benefit: at \rem{40} the training score of that variant, which is \sparse{} with the terminal score replaced by $\Phi(s_T)$, does not improve ($-0.13$ points per iteration, two-seed mean), where \civtelescope at every decision improves by $0.44$ (Appendix~\ref{app:controls}).

\paragraph{Ordering of hybrid rewards in \civmarshmallow.}
Reversing the order of the rewards within a game removes the hybrid rewards' gain.
The hybrid rewards of Section~\ref{sec:hybrid}, \scoreboard{}\hto\civtelescope and \sparse{}\hto\civtelescope, gain 5.21 and 5.15 points at \rem{120}.
\civtelescope{}\hto\sparse gains 1.76, and \scoreboard{}\hto\civtelescope{}\hto\sparse, which returns to \sparse at turn 80, gains 2.84 (Table~\ref{tab:ablation}).
\civtelescope should therefore take over after the early game and keep its weight to the end.

\section{Conclusion}

We show that current game scores frequently disagree with eventual outcomes, creating \textit{scoreboard traps} that are also difficult for off-the-shelf LLM judges to recognize. 
We introduce \civharness, which branches and replays saved game states to obtain grounded long-horizon feedback, and use these replay outcomes to train \civtelescope, a judge that more reliably distinguishes positions by their future value. 
We then use \civtelescope as a dense reward within \civmarshmallow, a phased training recipe for Freeciv agents. 
We show that replay-grounded rewards are especially useful in the middle of the game and \civtelescope can already rank positions, while combining them with simpler signals yields stronger full-game training. 
More broadly, our findings suggest that for long-horizon agents, useful reward signals need not come directly from what looks good now: grounding feedback in realized future outcomes can provide a better basis for learning gameplay. 

\section*{AI Use}
\label{sec:repro}

In this work, we used generative AI tools, primarily Codex and Claude Code, to implement methods, execute experiments, implement analyses selected by the authors, and parse experimental output logs. 
We have not used generative AI tools to develop research ideas or hypotheses or to formulate mathematical claims or proofs; the intellectual ideas and choice of analyses originated with the authors.
Additionally, we used generative AI tools to create and edit code and improve the paper’s readability. 
We did not use generative AI tools for literature searches. 
The use of language-model agents within gameplay simulations is part of the experimental methodology described in the paper. 
The authors manually verified all written text and citations and checked all experiments for correctness. 
We take responsibility for the final content of this work, including text, claims, and artifacts produced with the aid of generative AI.

\bibliography{references}
\bibliographystyle{iclr2027_conference}
\clearpage

\appendix
\addtocontents{toc}{\protect\setcounter{tocdepth}{2}}
\begingroup
\renewcommand{\contentsname}{Appendix Contents}
\tableofcontents
\endgroup
\clearpage
\section{Background}

\subsection{Freeciv rulesets}
\label{app:rulesets}
A Freeciv ruleset is a set of data files that defines the units, buildings, technologies, governments, terrain, and game parameters, so the same server plays different games under different rulesets.
We use two rulesets shipped with Freeciv 3.2.5.
\textsc{civ2civ3}, the default ruleset of the Freeciv 3 server, keeps the rules of Civilization II as its reference and moves the game toward Civilization III, aiming at slower and more linear city development; \textsc{classic} is the older standard ruleset from which it departs.
Relative to \textsc{classic}, \textsc{civ2civ3} removes rapture growth, so celebrating cities no longer grow every turn; makes Settlers cost two citizens instead of one; adds food upkeep for most units; enables plague in cities larger than four; makes a technology cheaper when a nation the player has an embassy with already knows it, and charges a per-city research upkeep late in the game; and requires at least three tiles between cities.
Scoreboard traps occur under both rulesets, in 26\% of decidable position pairs under \textsc{civ2civ3} and 27\% under \textsc{classic} (\Cref{fig:xgame_traps}(a)).
The main experiments, the judge's training data, and its held-out evaluation use \textsc{civ2civ3}; \textsc{classic} is used only to test the judge on unseen rules (Appendix~\ref{app:transfer}).

\subsection{Cross-game trap probe}
\label{app:xgame}

Every Freeciv trap rate and trap pair in the paper uses the terminal target of Section~\ref{sec:oracle}, the mean terminal score of the evaluated player; Table~\ref{tab:oracle_specs} lists the continuation policy, branch count, game length, and pair rule behind each.

\begin{table}[t]
\caption{The future outcome behind each Freeciv measurement.
Every row uses the evaluated player's terminal score: $K$ branches are played from the position to the configured final turn by the built-in bots at the \emph{normal} bot difficulty and averaged, or a policy rollout's own terminal score is taken.
The judge's held-out set is that of Tables~\ref{tab:trap_accuracy} and~\ref{tab:judge_comparison} (in-domain column). Cross-game pairs join positions of different games at most ten turns apart; a pair is decidable when its outcome gap exceeds twice the pooled standard error of the branch means, or, for single rollouts, is at least six points.}
\label{tab:oracle_specs}
\begin{center}
\scriptsize
\setlength{\tabcolsep}{3pt}
\begin{tabular}{llllll}
\toprule
Measurement & Positions & Played out by & $K$ & Game length & Decidable pairs \\
\midrule
Fig.~\ref{fig:xgame_traps}, \textsc{civ2civ3}; Fig.~\ref{fig:headline}a & 40 games, turns 10--115 & bots & 8 & 120 turns & cross-game, $>2$ SE \\
Fig.~\ref{fig:xgame_traps}, \textsc{classic} & 40 games, turns 10--115 & bots & 8 & 120 turns & cross-game, $>2$ SE \\
Ruleset transfer (Table~\ref{tab:judge_comparison}) & same games, turns 10--65 & bots & 8 & 120 turns & cross-game, $>2$ SE \\
Table~\ref{tab:trap_horizon} & 10 games per length & bots & 8 & 60, 70, 80 turns & cross-game, $>2$ SE \\
Judge held-out set & bot games; rollouts & bots; the rollout & 8; 1 & 70 turns & cross-game, $>2$ SE; $\geq 6$ \\
Table~\ref{tab:sibling_trap} & training rollouts & the rollout & 1 & 120 turns & same start and turn, $\geq 6$ \\
\bottomrule
\end{tabular}
\end{center}
\end{table}

The chess bank holds 541 positions from 80 played games, seven snapshots per game, each valued by 20 stochastic Stockfish replays to the end with the mean result for the side to move as the oracle and material balance in pawns as the visible proxy.
The Othello bank holds 420 positions from 70 games, six snapshots per game, each valued by 24 stochastic replays with the final disc differential as the oracle and the current disc differential as the proxy.
Pairs are built exactly as for Freeciv: across games, bounded move gap, decidable when the oracle gap exceeds twice the pooled standard error.
The zero-shot evaluation uses 300 trap and 300 non-trap pairs per game, both presentation orders, using the Freeciv prompt with only the game name changed.
Calls that time out are excluded rather than scored.
Table~\ref{tab:xgame_judges} reports the per-judge results.

\begin{table}[t]
\caption{Zero-shot accuracy of the five API judges on chess and Othello by pair slice.
Each game uses 300 trap and 300 non-trap pairs in both presentation orders, with the Freeciv prompt and only the game name changed; chance is 0.5.}
\label{tab:xgame_judges}
\begin{center}
\small
\begin{tabular}{lrrrr}
\toprule
 & \multicolumn{2}{c}{Chess} & \multicolumn{2}{c}{Othello} \\
\cmidrule(lr){2-3}\cmidrule(lr){4-5}
Judge & Non-trap & Trap & Non-trap & Trap \\
\midrule
GPT-5.4-mini     & 0.938 & 0.157 & 0.847 & 0.188 \\
DeepSeek-V3.2    & 0.862 & 0.315 & 0.815 & 0.263 \\
GPT-5.2          & 0.892 & 0.352 & 0.785 & 0.637 \\
Claude Sonnet 5  & 0.883 & 0.485 & 0.797 & 0.522 \\
GPT-5.6-sol      & 0.677 & 0.598 & 0.492 & 0.813 \\
\bottomrule
\end{tabular}
\end{center}
\end{table}

Five further banks (Table~\ref{tab:xgame_prevalence}) apply the same pipeline to the OpenSpiel implementations of backgammon, 9$\times$9 Go, Hearts, and 2048, and to the Catanatron implementation of Settlers of Catan.
Positions are snapshots of self-play by a deliberately weak stochastic engine, Monte Carlo tree search with 30 simulations of one random rollout each and 10\% uniformly random moves, which plays the role of the skill-level noise of Stockfish in chess.
2048 uses a one-ply greedy policy with the same random fraction and Hearts uniform random play, because tree search with terminal rewards applies neither to a single-player game with intermediate rewards nor to an imperfect-information game.
Each position is replayed $K{=}100$ times by the same engine ($K{=}48$ for Catan, where every snapshot yields four positions, one per seat, sharing the same continuations).
The visible proxy is the number the game itself displays: the pip-count lead in backgammon, the naive area count with komi in Go, penalty points relative to the table in Hearts, the running score in 2048, and public victory points in Catan, whose rules hide victory-point cards from the opponents so that the oracle, the same player's final victory points, includes what the proxy cannot see.
Where the game has a natural terminal quantity of the same kind, the oracle is that quantity (area margin, penalty, score, victory points); in backgammon it is the mean result.
Every bank, pair set, and engine setting is reproducible from the repository.
For the progress axis of Figure~\ref{fig:xgame_traps}b a pair is placed at the mean position of its two snapshots divided by the typical length of a complete game under the same engine (last row of the figure), and bins with fewer than 30 pairs are left blank.
Table~\ref{tab:xgame_progress} lists the rate in each of the ten progress bins.
The Freeciv rows use the first two banks of Table~\ref{tab:oracle_specs}, built with one protocol: 40 games of 120 turns per ruleset, labeled every five turns from turn 10 to 115, 40 new \textsc{civ2civ3} games (57,846 decidable pairs with untied scores) and the 40 \textsc{classic} games of the ruleset transfer (55,873 such pairs).

\begin{table}[t]
\caption{Trap prevalence in every cross-game bank.
Positions are snapshots of self-play games between copies of the named engine (80 chess, 70 Othello, 40 backgammon, 59 Go, 60 Hearts, 60 2048, and 40 Catan games); the oracle replays each position $K$ times to the end with the same engine and takes the mean outcome.
Pairs are built as in Freeciv: across games, bounded move gap, decidable when the oracle gap exceeds twice its pooled standard error; the trap rate is taken over decidable pairs whose visible proxy is not tied.
Zero-shot judges are evaluated on the chess and Othello banks only (Table~\ref{tab:xgame_judges}).}
\label{tab:xgame_prevalence}
\begin{center}
\scriptsize
\setlength{\tabcolsep}{4pt}
\begin{tabular}{lllrrrr}
\toprule
Game (genre) & Visible proxy & Engine & Positions & $K$ & Pairs & Trap rate \\
\midrule
Chess (board) & material balance & Stockfish, skill-level noise & 541 & 20 & 34,024 & 16.0\% \\
Othello (board) & disc differential & depth-2 search + temperature & 420 & 24 & 18,334 & 68.2\% \\
Backgammon (board, dice) & pip-count lead & MCTS-30 + 10\% random & 187 & 100 & 6,587 & 8.1\% \\
Go (9x9) (board) & area count with komi & MCTS-30 + 10\% random & 367 & 100 & 6,182 & 38.8\% \\
Hearts (card) & penalty so far vs.\ table & uniform random & 300 & 100 & 14,543 & 17.7\% \\
2048 (single-player puzzle) & score & one-ply greedy + 10\% random & 381 & 100 & 12,042 & 31.2\% \\
Catan (4-player economic) & public victory points & victory-point greedy bots & 952 & 48 & 79,737 & 14.4\% \\
\bottomrule
\end{tabular}
\end{center}
\end{table}

\begin{table}[t]
\caption{Trap rate (\%) in each progress bin of Figure~\ref{fig:xgame_traps}b.
Progress is the mean position of a pair divided by the typical length of a complete game; a dash marks a bin with fewer than 30 pairs: no points are scored while cards are passed in Hearts, and the Freeciv banks cover turns 10--115 of 120-turn games (Table~\ref{tab:oracle_specs}).}
\label{tab:xgame_progress}
\begin{center}
\footnotesize
\setlength{\tabcolsep}{4pt}
\begin{tabular}{lrrrrrrrrrr}
\toprule
Progress (\%) & 0--10 & 10--20 & 20--30 & 30--40 & 40--50 & 50--60 & 60--70 & 70--80 & 80--90 & 90--100 \\
\midrule
Backgammon & 10 & 5 & 6 & 8 & 9 & 10 & 9 & 11 & 9 & 6 \\
Catan & 25 & 16 & 15 & 16 & 12 & 11 & 11 & 12 & 13 & 12 \\
Chess & 6 & 19 & 21 & 17 & 12 & 9 & 5 & 3 & 4 & 2 \\
Hearts & -- & 18 & 20 & 23 & 22 & 18 & 12 & 8 & 7 & 4 \\
Freeciv (civ2civ3) & 55 & 49 & 40 & 35 & 29 & 28 & 24 & 18 & 10 & 8 \\
Freeciv (classic) & 41 & 46 & 46 & 35 & 31 & 28 & 24 & 17 & 11 & 7 \\
2048 & 73 & 37 & 38 & 38 & 29 & 30 & 41 & 32 & 24 & 19 \\
Go (9x9) & 0 & 13 & 21 & 45 & 58 & 46 & 37 & 30 & 21 & 15 \\
Othello & 61 & 50 & 72 & 72 & 69 & 65 & 59 & 47 & 43 & 33 \\
\bottomrule
\end{tabular}
\end{center}
\end{table}

\subsection{Game-length curve}

For each game length in Table~\ref{tab:trap_horizon} we generated ten fresh small-map games from disjoint seed pools and labeled every fifth turn with the replay oracle at $K{=}8$ branches at the \emph{normal} bot difficulty.
Per turn bucket we correlate the current scoreboard and the oracle value with the actual terminal score of the real game.
At 80 turns the oracle's own early-bucket correlation is 0.10--0.39, so early positions in long games need more branches or labels built phase by phase; the training runs use reference sets rebuilt for the phase being trained for this reason.

\begin{table}[t]
\caption{%
The scoreboard predicts the terminal score less well in longer games.
For ten games at each game length, we correlate the current scoreboard and the replay-oracle value ($K{=}8$ branches) with the game's terminal score at every fifth turn, and average the correlations over turns.
``Turns won'' counts the sampled turns at which the oracle value correlates at least as well as the scoreboard; the trap rate is the share of decidable position pairs whose scoreboard order disagrees with their terminal-score order.}
\label{tab:trap_horizon}
\begin{center}
\small
\begin{tabular}{lrrrr}
\toprule
Game length & Scoreboard corr. & Oracle corr. & Turns won by oracle & Trap rate \\
\midrule
60 turns & 0.81 & 0.98 & 7 / 10 & 30\% \\
70 turns & 0.53 & 0.95 & 11 / 12 & 31\% \\
80 turns & 0.12 & 0.58 & 12 / 14 & 54\% \\
\bottomrule
\end{tabular}
\end{center}
\end{table}

\subsection{Within-game trap rate}
\label{app:sibling}

The cross-game pair rule of Section~\ref{sec:oracle} keeps the turn number from leaking into the comparison, but it raises the question of whether the trap merely reflects incomparable maps.
We therefore measure the trap rate on pairs of policy rollouts that share the same start, that is the same map, seed, opponents, and save, and are compared at the same turn, so that the two positions differ only in what the focal policy did.
The rollouts are the archived training games of 16 arms from the scoreboard, judge, and hybrid families in every training phase and seed, forming 640 arm-by-start clusters, with games that end in elimination excluded.
A pair counts when the realized terminal scores differ by at least six points, the rule used for the judge's policy-rollout training pairs; a trap is a pair whose leader on the visible score finishes behind, and pairs with tied visible scores are dropped.
Cross-start pairs from the same rollouts at the same turn give the comparison.
Table~\ref{tab:sibling_trap} reports both rates by turn bucket.
Same-start pairs are traps 18--23\% of the time through turn 69 and 7--10\% afterwards, and from turn 30 on they are traps two to ten times more often than cross-start pairs, because start luck no longer explains the score gap.
The trap is therefore not an artifact of cross-game pairing.

\begin{table}[t]
\caption{Within-game trap rate on the training rollouts.
Pairs of rollouts from the same start, compared at the same turn, whose realized terminal scores differ by at least six points; a trap is a pair whose leader on the visible score finishes behind, with tied visible scores dropped.
Cross-start pairs are drawn from the same rollouts at the same turn under the same rule.
Intervals are cluster bootstraps over 640 arm-by-start clusters.}
\label{tab:sibling_trap}
\begin{center}
\small
\begin{tabular}{lllrr}
\toprule
Turn bucket & Same-start trap rate & 95\% interval & Same-start pairs & Cross-start trap rate (pairs) \\
\midrule
10--29 & 0.233 & [0.173, 0.304] & 7,255 & 0.298 (95,108) \\
30--49 & 0.179 & [0.152, 0.208] & 31,384 & 0.113 (216,768) \\
50--69 & 0.219 & [0.199, 0.241] & 104,653 & 0.037 (327,427) \\
70--89 & 0.102 & [0.091, 0.114] & 119,881 & 0.019 (348,151) \\
90--119 & 0.071 & [0.063, 0.079] & 389,279 & 0.010 (742,316) \\
\bottomrule
\end{tabular}
\end{center}
\end{table}

\section{\civharness}
\label{app:details}

\subsection{Harness acceptance tests}
\label{app:harness}

The harness acceptance suite checks replica identity, deterministic re-seeded branching, a known-order positive control, branch-level noise floors, turn-matched readouts, faithful client actions, restricted observation/action spaces, and complete game replay.
A branch is excluded from an analysis if its provenance digest or restored opponent configuration fails validation.

Replaying the same save, branch seed, and action trace must reproduce the same state digest.
An \texttt{EnvSpec} separates the raw engine state from the observation and action space exposed to an agent, allowing selected economic, technology, diplomacy, unit, and terrain fields to be exposed without access to the raw save.
Game logs include the observation digest and restored bot difficulty so that a stochastic language-model run remains replayable.

\section{Training a Freeciv gameplaying agent}

\subsection{Judge training and calibration}
\label{app:judge}

The judge is evaluated on 600 non-trap and 600 trap pairs from games and rollouts excluded from training.
For each sampled game turn of the phase being trained, the reference set of Equation~\ref{eq:potential} is drawn from eight positions with known terminal scores: six from policy rollouts, evenly spaced in rank by terminal score, and the two highest-scoring bot-played positions.
During training $\mathcal{B}$ holds four of the eight, chosen to span the score range ($M{=}4$), and the judge is queried in one presentation order.
In the \judge{} and hybrid arms a candidate set is rebuilt every $N{=}4$ policy updates from 40 recent training rollouts and validated on the 40 rollouts that follow them.
The candidate replaces the set in use only if the Spearman correlation between its potential and the terminal score is at least $\rho_{\min}{=}0.2$ both at turn 40 and at the last decision (Equation~\ref{eq:reference_update}); the set never changes within a policy update.
The potential is sensitive to the score range of the reference set.
In the \rem{120} phase of seed 44, replacing only the reference set, from one whose four scores ranged from 7 to 33 to one ranging from 2 to 36, turned the training-score slope from $-0.06$ to $+0.06$ points per iteration; on seed 43 the slope was positive with both sets ($0.10$ and $0.16$).

On the terminal-score label set, full-parameter SFT memorizes the training set and lands at chance on held-out pairs while dismantling the base model's score-reading prior; LoRA with rank 16 on the same data, labels, and prompt generalizes, and it is stable across three training seeds.
Adding more pairs beyond about seven thousand does not raise held-out accuracy.

\subsection{Hybrid schedules at \rem{120}}
\label{app:hybrid}

\scoreboard{}\hto\civtelescope, \sparse{}\hto\civtelescope, and \civtelescope{}\hto\sparse (Section~\ref{sec:hybrid}) use transition turn $\tau=35$ and width $\beta=5$, so the judge weight is $w(u)=(1+\exp(-(u-35)/5))^{-1}$.
\scoreboard{}\hto\civtelescope{}\hto\sparse first blends the \scoreboard and \civtelescope potentials with $w(u_t)$, then shifts from the resulting dense advantage to the sparse advantage using a second sigmoid centered at turn 80 with scale 5.

Tables~\ref{tab:phase_diagram} and~\ref{tab:ablation} list the \rem{120} phase gain of every schedule.
Every hybrid and the pure judge arm start from the judge $\phend[80]$ checkpoint of their seed; the pure sparse and scoreboard arms start from their own $\phend[80]$ checkpoints.
\civtelescope{}\hto\sparse was replicated in two seeds, the others in three.
All eleven hybrid $\phend[120]$ checkpoints have also been rolled out on the \rem{80} evaluation set $\evalset[80]$: \scoreboard{}\hto\civtelescope scores 58.4, 57.0, and 56.7 for seeds 43, 44, and 45; \sparse{}\hto\civtelescope 58.2, 59.3, and 51.9; \civtelescope{}\hto\sparse 58.7 and 54.4 for seeds 43 and 44; and \scoreboard{}\hto\civtelescope{}\hto\sparse 59.6, 59.0, and 51.8.
The base model scores 40.7 on this set and the single-signal $\phend[80]$ checkpoints 52.4--60.2.
The spread across seeds follows the $\phstart[120]$ checkpoint: seed 45's judge $\phend[80]$ checkpoint, from which its hybrids start, scores 52.4 on this set, and its three hybrids score 51.8--56.7.

\section{Experiments}

\subsection{Aggregation}

For each starting position, the eight games are averaged before computing a phase gain, and a game in which the agent's player is eliminated counts as a score of zero.
Paired win rates compare these start-level gains.

\subsection{Judge evaluation and zero-shot judges}
\label{app:zero_shot}

Off-the-shelf models are scored on a seeded subsample of 300 trap and 300 non-trap pairs of the judge's held-out set, with the same fog renderings and prompt as the judge; intervals are cluster bootstraps over source games, and ratios are paired over the same pairs (Table~\ref{tab:trap_accuracy}).
The judge's accuracies on the full 600-pair slices are 0.778 and 0.866.
Every pair is presented in both orders and accuracy is content-level.

\begin{table}[t]
\caption[Trap accuracy on the judge's held-out set.]{Trap accuracy on the judge's held-out set: 300 trap and 300 non-trap pairs from games and rollouts excluded from training, both presentation orders, content-level.
Chance is 0.5.
\emph{Warned} repeats the trap pairs with a sentence in the prompt saying that the visible score can mislead and that the future outcome decides.
The last column is how many times more accurate the judge is than the model on the same trap pairs, with a 95\% paired bootstrap interval clustered over source games.}
\label{tab:trap_accuracy}
\begin{center}
\small
\setlength{\tabcolsep}{4pt}
\begin{tabular}{lrrrl}
\toprule
Judge & Non-trap pairs & Trap pairs & Warned & Times more accurate \\
\midrule
Qwen2.5-7B-Instruct (zero-shot backbone) & 0.766 & 0.260 & --- & 1.98 [1.22, 3.27] \\
GPT-5.6-luna     & 0.792 & 0.258 & 0.317 & 2.00 [1.27, 3.25] \\
GPT-5.5          & 0.805 & 0.262 & 0.347 & 1.96 [1.25, 3.10] \\
GPT-5.6-terra    & 0.823 & 0.293 & 0.355 & 1.64 [1.03, 2.64] \\
GPT-5.6-sol      & 0.802 & 0.330 & 0.378 & 1.35 [0.87, 2.13] \\
GPT-6-astra      & 0.818 & 0.343 & 0.378 & 1.26 [0.76, 2.03] \\
\midrule
Replay-grounded judge (same 7B base) & \textbf{0.878} & \textbf{0.775} & --- & --- \\
\bottomrule
\end{tabular}
\end{center}
\end{table}

\subsection{Regression value head protocol}
\label{app:value_head}

The head shares \civtelescope's base, adapter, and recipe (Qwen2.5-7B, LoRA rank 16, two epochs, one seed) and replaces the pairwise objective with a scalar output trained to predict a position's replay outcome, the mean terminal score of its $K{=}8$ branches, under a Huber loss on standardized targets.
Per-position replay outcomes are available for 2,900 positions of the judge's training bank; the head is trained on the 2,583 of them from games outside the held-out set and read after the two epochs, without snapshot selection.
On a pair it picks the position with the higher predicted score.
It is scored on the 200 held-out trap pairs and 197 non-trap pairs whose positions carry a per-position label, and \civtelescope and the zero-shot backbone are re-read on the identical pairs from their existing answers: the head scores 0.520 $[0.317, 0.688]$ on trap pairs and 0.548 on non-trap pairs, \civtelescope 0.688 and 0.777, and the backbone 0.219 and 0.612.
The head fits its training positions (Pearson $r = 0.77$ with the replay outcome) but not held-out positions ($r = 0.16$).
On the classic pairs of Appendix~\ref{app:transfer}, 1,800 trap and 600 non-trap, the head scores 0.622 $[0.481, 0.750]$ and 0.703, \civtelescope 0.592 $[0.515, 0.672]$ and 0.741, and the backbone 0.241 and 0.818.
The head has no presentation order and gives one pick per pair, while \civtelescope keeps its content-level score over both orders.
No classic game appears among the head's training games, and its predictions correlate weakly with the turn ($r = 0.15$) and with the visible score ($r = 0.24$).

\subsection{Ruleset transfer protocol}
\label{app:transfer}

The classic bank holds 40 games generated under the classic ruleset with the judge's own protocol: positions every five turns from turn 10 to 65 of 120-turn games, 480 positions in all, each labeled by $K{=}8$ branches replayed to the end of the game at the \emph{normal} bot difficulty and rendered from the focal player's view.
Pairs follow the training rule (cross-game, $|\Delta\text{turn}|\leq 10$, decidable at two pooled standard errors), giving 26,819 decidable pairs of which 39.1\% are traps; the evaluation set is a seeded subsample of 1,800 trap and 600 non-trap pairs, and five frontier models are scored on 200 trap and 100 non-trap pairs of it.
The readouts were frozen before judging: the paired trap-accuracy difference against the zero-shot backbone, $+0.351$ $[0.248, 0.454]$ over 10,000 bootstrap draws clustered by game, and against the strongest zero-shot judge on the API subsample, with the strongest re-taken inside each draw, $+0.31$ $[0.152, 0.421]$ over the backbone and GPT-5.6-sol as frozen and $+0.07$ $[-0.08, 0.21]$ once the four further frontier models are included.
The non-trap difference is $-0.077$ $[-0.188, 0.018]$.

\subsection{Offline window diagnostic}
\label{sec:diagnostic}

Equation~\ref{eq:gae} consumes differences of $\Phi$ between successive decisions, so a potential helps in proportion to how well it \emph{ranks} states, not to how small its pointwise error is.
Table~\ref{tab:knowability_spectrum} buckets the decisions of the \rem{120} rollouts of two seeds by turn and reports, per bucket, the median absolute error and the Spearman correlation of the judge potential and of the scoreboard with the realized terminal score.
The judge's pointwise error is half the scoreboard's from the first bucket, yet its rank correlation is only 0.09 there; it reaches 0.54--0.60 at turns 50--69, within the \rem{80} phase that begins at turn 40.
The scoreboard's rank correlation exceeds the judge's in every bucket, because the quantities that make up the current score persist into the terminal score, and reaches 0.98 in the last bucket; the judge's explained variance beyond the scoreboard is largest at \rem{80}, 24--26\% against 2--11\% in the other phases.

\begin{table}[t]
\caption{Offline window diagnostic on \rem{120} rollouts.
Per turn bucket, the median absolute error and the Spearman correlation of each potential against the realized terminal score, for two training seeds (a / b).
The judge's error is below the scoreboard's from the first bucket, but its rank correlation only exceeds 0.4 from the 30--49 bucket onward, and the scoreboard becomes an identity in the last bucket.}
\label{tab:knowability_spectrum}
\begin{center}
\small
\begin{tabular}{lrrrr}
\toprule
 & \multicolumn{2}{c}{Replay-grounded judge} & \multicolumn{2}{c}{Scoreboard} \\
\cmidrule(lr){2-3}\cmidrule(lr){4-5}
Turn bucket & Median error & Spearman $\rho$ & Median error & Spearman $\rho$ \\
\midrule
10--29   & 7.8 / 9.6 & 0.09 / 0.10 & 15.0 / 16.0 & 0.34 / 0.36 \\
30--49   & 4.8 / 6.2 & 0.42 / 0.41 & 12.0 / 12.0 & 0.64 / 0.65 \\
50--69   & 3.9 / 5.1 & 0.60 / 0.54 & 8.0 / 8.0   & 0.78 / 0.78 \\
70--89   & 3.6 / 3.9 & 0.64 / 0.65 & 6.0 / 6.0   & 0.86 / 0.87 \\
90--109  & 2.9 / 3.0 & 0.74 / 0.76 & 3.0 / 3.0   & 0.92 / 0.92 \\
110--129 & 2.7 / 3.2 & 0.77 / 0.79 & 0.0 / 0.0   & 0.98 / 0.98 \\
\bottomrule
\end{tabular}
\end{center}
\end{table}

\subsection{Prompting-scaffold panel}
\label{app:scaffolds}

The five prompting-based agents in Table~\ref{tab:scaffold_panel} are direct prompting; BaseLang, which writes a chain of reasoning before the action, and Mastaba, a hierarchical design with one summarizing call over several advisor calls, both from CivRealm \citep{qi2024civrealm}; the scene-graph planner SAGA \citep{li2026saga}; and the self-reflection loop Reflexion \citep{shinn2023reflexion}.
We re-implement the four published designs on our menu interface with our own prompts.
Of the 1,800 prompting games, 1,684 completed and the rest failed on infrastructure.
All use the same Qwen3-8B base, the same menu interface, and the same opponents as the trained policies, on the \rem{120} evaluation set $\evalset[120]$.
The trained policies are the checkpoints produced by the \rem{120} phase, the last of the four training phases, on that base.
Prompt cost dominates the per-call cost, so the trained policy and the scaffolds with about one call per decision cost the same; only the hierarchical advisor scaffold, with four calls per decision, is materially more expensive.

\subsection{Label-source ablation protocol}
\label{app:label_source}

The two judges are trained on the same 3,592 pairs, a seeded subset of the judge's training pairs, with the judge's recipe (Qwen2.5-7B, LoRA rank 16, two epochs, one seed); one arm keeps the replay label of each pair, the other replaces it with GPT-5.6-sol's answer on that pair in one presentation order.
The labeler agrees with the replay outcome on 0.678 of the pairs overall and on 0.489 of the trap pairs.
Both arms are read at the training amount of the full judge (its 4,800 steps scaled to the subset) on the judge's held-out set, 600 trap and 600 non-trap pairs from games excluded from training.
At that point the replay arm scores 0.610 $[0.481, 0.677]$ on trap pairs and 0.859 on non-trap pairs; the preference arm scores 0.391 $[0.239, 0.482]$ and 0.763.
At each arm's best trap-accuracy snapshot the gap is the same, 0.692 against 0.444, and the two intervals are disjoint.
On the classic trap pairs of Appendix~\ref{app:transfer} these snapshots score 0.531 and 0.415, against 0.243 for the backbone.

\subsection{Additional mechanism controls}
\label{app:controls}

In the terminal-judge control the judge's potential of the final position replaces the terminal score, and everything else follows \sparse{}.
Its readout is the improvement in training score per iteration, estimated for each starting position over the twelve iterations of a training phase and averaged over positions.
On seeds 43 and 44 the control changes by $-0.19$ and $-0.07$ points per iteration and the judge applied at every decision by $+0.65$ and $+0.23$; \sparse{} improves by $1.13$ and $0.80$ and \scoreboard{} by $0.80$ and $0.19$.
The control starts \rem{40} from the base model, whereas the three signal families continue from their $\phend[20]$ checkpoints, and it uses a fixed reference set.
At \rem{80} its training score recovers from its weaker $\phend[40]$ checkpoint but still ends below \sparse{} on both seeds.

\end{document}